\pdfoutput=1

\documentclass[11pt]{article}

\usepackage[]{acl}

\usepackage{times}
\usepackage{latexsym}

\usepackage[T1]{fontenc}

\usepackage[utf8]{inputenc}

\usepackage{microtype}

\usepackage{inconsolata}

\usepackage{booktabs}
\usepackage{amsmath}
\usepackage{enumitem}
\usepackage{graphicx}
\usepackage{subcaption}
\usepackage{multirow}

\title{Unsupervised Multilingual Dense Retrieval via Generative Pseudo Labeling}

\author{Chao-Wei Huang$^{\dag\ddag}$\quad Chen-An Li$^\dag$\quad Tsu-Yuan Hsu$^\dag$ \quad Chen-Yu Hsu$^\dag$\quad Yun-Nung Chen$^\dag$\\
  $^{\dag}$National Taiwan University, Taipei, Taiwan\\
  $^{\ddag}$Taiwan AI Labs, Taipei, Taiwan \\
  \texttt{f07922069@csie.ntu.edu.tw\quad y.v.chen@ieee.org} \\}

\begin{document}
\maketitle
\begin{abstract}
Dense retrieval methods have demonstrated promising performance in multilingual information retrieval, where queries and documents can be in different languages. However, dense retrievers typically require a substantial amount of paired data, which poses even greater challenges in multilingual scenarios. This paper introduces \textbf{UMR}, an \underline{U}nsupervised \underline{M}ultilingual dense \underline{R}etriever trained without any paired data. Our approach leverages the sequence likelihood estimation capabilities of multilingual language models to acquire pseudo labels for training dense retrievers. We propose a two-stage framework which iteratively improves the performance of multilingual dense retrievers. Experimental results on two benchmark datasets show that UMR outperforms supervised baselines, showcasing the potential of training multilingual retrievers without paired data, thereby enhancing their practicality.\footnote{Our source code, data, and models are publicly available: \url{https://github.com/MiuLab/UMR}}

\end{abstract}

\section{Introduction}
\begin{figure}[t]
    \centering
    \includegraphics[width=\linewidth]{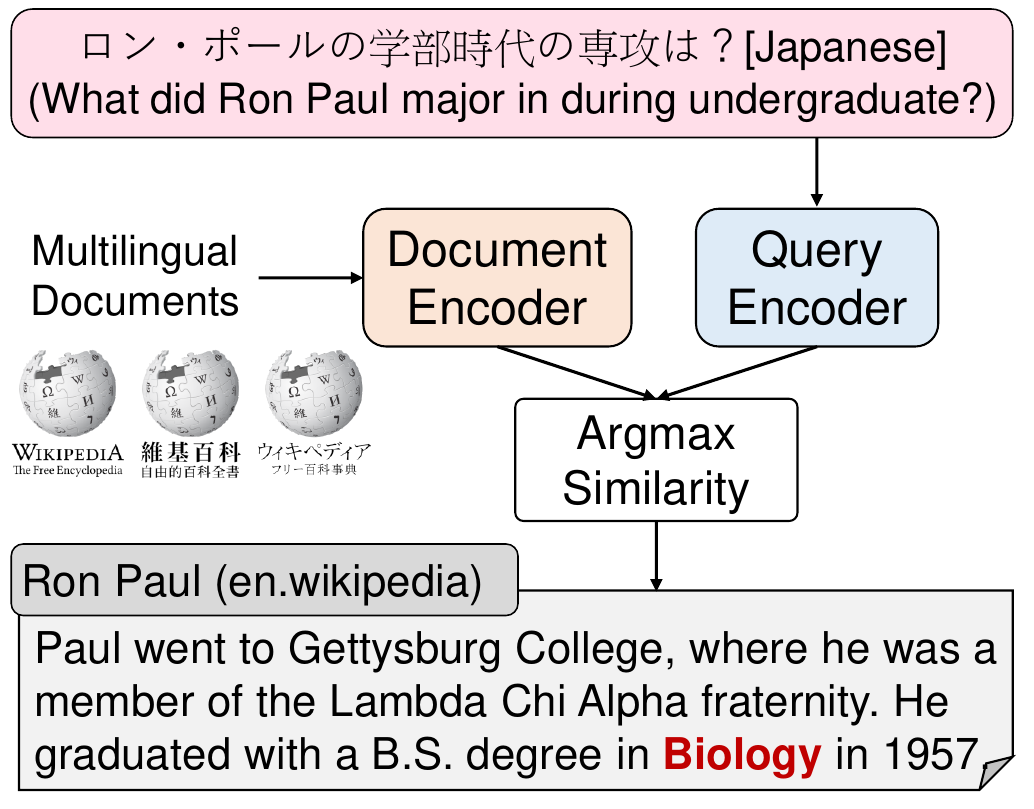}
    \caption{Illustration of the multilingual dense retrieval process. Given a query, the goal is to retrieve relevant documents in any language. Dense retrieval achieves this by encoding the query and documents into dense representations and performing vector similarity search.}
    \label{fig:mir}
\end{figure}

Multilingual information retrieval (mIR) has attracted significant research interest as it enables unified knowledge access across diverse languages. The task involves retrieving relevant documents from a multilingual collection given a query, which may be in a different language. Traditional sparse retrieval methods that rely on lexical matching often yield inferior performance due to the different scripts used~\cite{asai2021one}.
On the other hand, dense retrieval methods have shown promising results in multilingual retrieval by capturing semantic relationships between queries and documents~\cite{shen-etal-2022-recovering,zhang2022towards,ren-etal-2022-empowering,sorokin-etal-2022-ask}.
Figure~\ref{fig:mir} illustrates the process of multilingual dense retrieval.

Nevertheless, training dense retrievers requires a large amount of paired data, which is costly and time-consuming to collect. This challenge is particularly pronounced for low-resource languages where the availability of annotated data is limited. Consequently, there is a growing demand for more efficient techniques to build multilingual dense retrievers, such as leveraging unsupervised learning and transfer learning, to alleviate the data requirement.

The advance of large-scale language model pre-training~\cite{devlin-etal-2019-bert,conneau-etal-2020-unsupervised} presents a compelling avenue to explore, namely leveraging the multilingual capabilities of pre-trained multilingual language models. In this paper, we propose \textbf{UMR}, an unsupervised approach to multilingual dense retrieval that only relies on multilingual queries \emph{without requiring any paired data}. Our method leverages the sequence likelihood estimation capabilities of multilingual language models to obtain pseudo labels by estimating the conditional probability of generating the query given the document. This allows training of multilingual dense retrievers in a fully unsupervised manner.

To evaluate the effectiveness of our approach, we conduct experiments on XOR-TyDi QA~\cite{asai-etal-2021-xor}, a widely used benchmark for multilingual information retrieval. Our results demonstrate that \textbf{UMR} outperforms or performs comparably to existing supervised baselines on both XOR-Retrieve and XOR-Full. Additionally, we conduct comprehensive ablation studies to analyze the impact of different components of our approach. Our approach shows great potential for being applied to a broad range of multilingual information retrieval tasks, where it can reduce the dependence on costly paired data.

Our contributions can be summarized in 3-fold:
\begin{itemize}
    \item We propose \textbf{UMR}, the first unsupervised method for training multilingual dense retrievers without any paired data.
    \item Experimental results on two benchmark datasets show that our proposed method performs comparable to or even outperforms strong supervised baselines.
    \item The detailed analysis justifies the effectiveness of individual components in our \textbf{UMR}.
\end{itemize}

\section{Related Work}

\paragraph{Dense Retrieval}
Dense retrieval has garnered significant attention for its potential to enable retrieval in the semantic space. A prominent method in this area is the dense passage retriever (DPR)~\cite{karpukhin-etal-2020-dense}, which comprises a query encoder and a passage encoder. Several studies have also explored efficient training approaches, such as RocketQA~\cite{qu-etal-2021-rocketqa} and alternative architectures for dense retrieval, e.g., ColBERT~\cite{khattab2020colbert}.
A common technique for training performant dense retrievers is knowledge distillation from cross encoders. BERT-CAT~\cite{hofstaetter2020_crossarchitecture_kd} proposed cross-architecture knowledge distillation to improve dense retrievers and rankers. ~\citeauthor{izacard2020distilling} distilled knowledge from the reader model to the retriever model, thus improving its performance on open-domain question answering.
However, the majority of previous work has primarily focused on English retrieval, limiting its applicability to other languages.

\paragraph{Multilingual Dense Retrieval}
Multilingual information retrieval has been an active research area for several decades. Early work in this field primarily focused on cross-lingual information retrieval (CLIR), aiming to retrieve relevant documents in a different language from the query language~\cite{nasharuddin2010cross}.
Traditional CLIR systems relied on aligning bilingual dictionaries or parallel corpora to translate queries or documents into a common language for retrieval. However, these systems often faced limitations in translation quality, vocabulary coverage, and handling domain-specific expressions~\cite{ballesteros1996dictionary,vulic2015monolingual,sharma2016cross}.

In recent years, dense retrieval has emerged as a promising approach for multilingual information retrieval. Various studies have demonstrated the effectiveness of dense retrieval methods in cross-lingual and multilingual scenarios. Models such as XLM-R~\cite{conneau-etal-2020-unsupervised} and mBERT~\cite{devlin-etal-2019-bert} have achieved remarkable performance on diverse natural language processing tasks, including similarity-based retrieval tasks. The success of these models has spurred researchers to explore their application in multilingual information retrieval~\cite{jiang2020cross}.

\paragraph{Supervised mIR}
Most existing multilingual retrieval models rely on supervised training, where paired data consisting of queries and corresponding relevant documents in different languages is required.
These methods use popular datasets such as Mr. TyDi~\cite{zhang-etal-2021-mr} and XOR-TYDI QA~\cite{asai-etal-2021-xor}.
DR.DECR proposes to leverage the knowledge of an English retriever to improve cross-lingual retrieval~\cite{li-etal-2022-learning-cross}.
It uses paired data for machine translation to align multilingual representations.
Quick proposes to leverage supervised question generation to improve cross-lingual dense retrieval~\cite{ren-etal-2022-empowering}.
However, these methods still rely on question-document pairs and paired translation data.
The requirement for paired training data can be a significant bottleneck for multilingual information retrieval, especially for low-resource languages, where it is challenging to obtain large amounts of data.
In contrast, our method does not require any paired data or paired translation data, eliminating the requirement for annotation resources.

\paragraph{Unsupervised Dense Retrieval}
There have been recent efforts to develop unsupervised or weakly supervised approaches to dense retrieval.
InPars~\cite{bonifacio2022inpars}, Promptagator~\cite{dai2022promptagator}, and CONVERSER~\cite{huang-etal-2023-converser} 
 all proposed to generate synthetic queries with LLMs from few-shot examples, which achieved comparable performance to supervised methods in dense retrieval. However, synthetic query generation is less suitable for the multilingual setting as multilingual query generation remains a hard problem for multilingaul LLMs, which is demonstrated in our experiments.
UPR and ART are the most closely related work to our work~\cite{sachan-etal-2022-improving,sachan2022questions}.
UPR proposes to rerank passages with zero-shot question generation, which only requires a base LLM.
ART proposes to train a retriever without paired data with unsupervised reranking by language models.
Our method is similar to the framework proposed in ART, while we focus on multilingual scenarios where supervised data is even harder to collect.

\begin{figure*}[t]
    \centering
    \begin{subfigure}{\textwidth}
        \includegraphics[width=\textwidth]{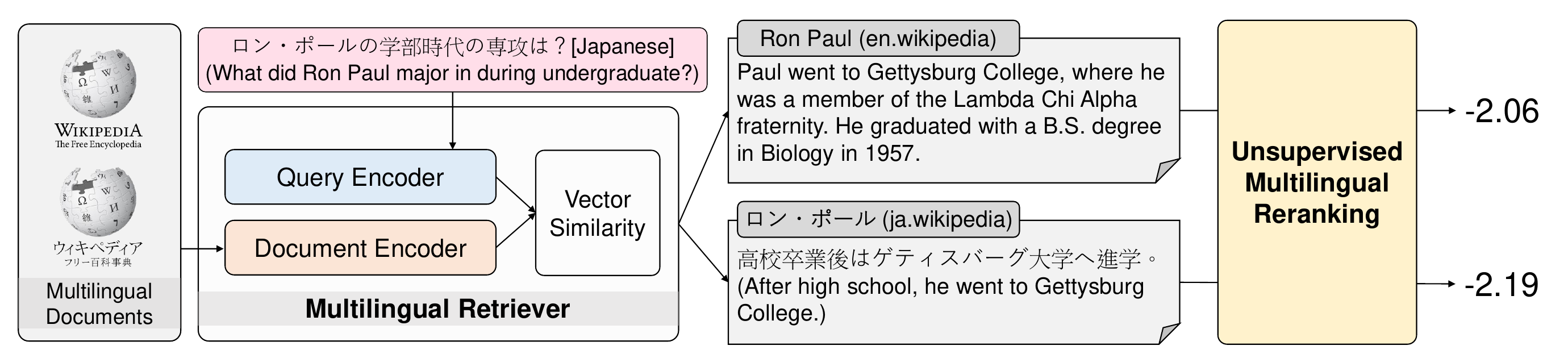}
        \caption{Stage 1: unsupervised multilingual reranking.}
        \label{fig:reranking_diagram}
    \end{subfigure}
    \vspace{3mm}
    \begin{subfigure}{\textwidth}
        \includegraphics[width=\textwidth]{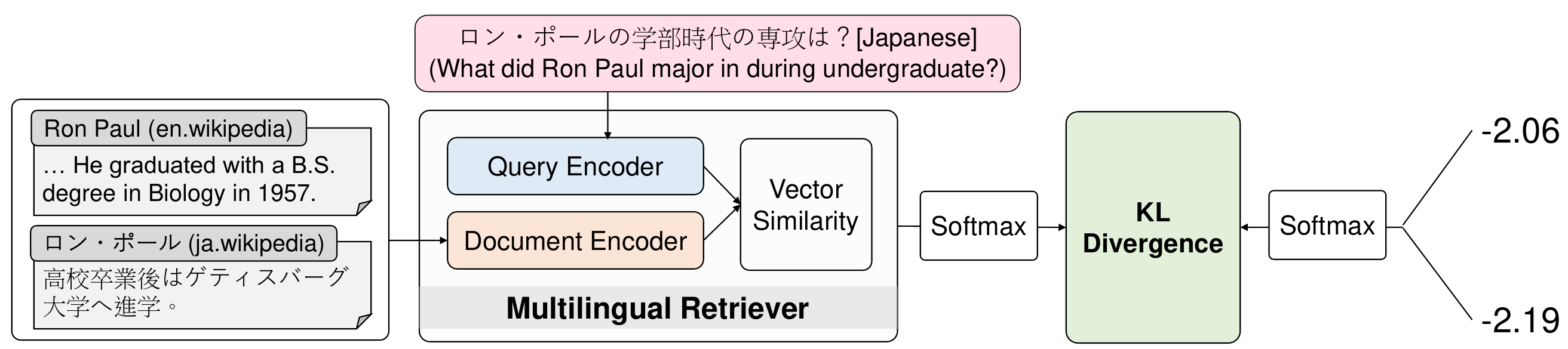}
        \caption{Stage 2: knowledge-distilled retriever training.}
        \label{fig:distillation_diagram}
    \end{subfigure}
    \caption{Illustration of our proposed UMR, unsupervised multilingual dense retrieval.}
    \label{fig:framework}
\end{figure*}

\paragraph{Multilingual Evidence for Fact Checking}
The power of generative models has made it easier for misleading information to spread, posing challenges in its detection~\cite{shu2017fake,wang2017liar}.
Previous fact-checking research has considered single-language evidence, often lacking sufficient cues for verification.
\citet{dementieva2023multiverse} proposed the use of multilingual evidence as features for fake news detection, resulting in improved performance. While our method does not specifically focus on fact checking, it can be applied to assist in finding multilingual evidence, thereby enhancing the verification process.

In this paper, we introduce an unsupervised multilingual dense retrieval approach that leverages the generative capabilities of multilingual language models to obtain pseudo labels for training the dense retriever.
Our method eliminates the need for paired training data, making it particularly suitable for low-resource languages.

\section{Our Method: UMR}
The goal of multilingual information retrieval is to retrieve relevant documents, denoted as $D^{+}$, from a collection of multilingual documents $\mathcal{D} = {d_1, \cdots, d_n}$. We adopt a widely used dense retrieval architecture, DPR~\cite{karpukhin-etal-2020-dense}, comprising a query encoder $E_{q}$ and a document encoder $E_{d}$. The documents are pre-encoded using the document encoder and then indexed for efficient vector search. Given a query $q$, the relevance score of a query-document pair is computed as their vector similarity:
\begin{equation*}
    r(q, d_i) = E_{q}(q)^\top E_{d}(d_i)
\end{equation*}

This section introduces our proposed framework \textbf{UMR} for training unsupervised multilingual retrievers iteratively. 
The framework consists of two stages: 1) unsupervised multilingual reranking and 2) knowledge-distilled retriever training, as illustrated in Figure~\ref{fig:framework}.

\subsection{Unsupervised Multilingual Reranking}

In the first stage, we leverage the generative capabilities of multilingual language models to rerank retrieved passages and obtain pseudo labels for training the dense retriever in an unsupervised manner. This stage is depicted in Figure~\ref{fig:reranking_diagram}.

Formally, given a query $q$ in language $L$, we retrieve the top-k documents ${d_1, \cdots, d_k}$ from the multilingual document collection using a multilingual dense retriever, forming $k$ query-document pairs.
We then utilize a pre-trained autoregressive multilingual language model (mLM) for unsupervised multilingual reranking.
For each query-document pair $(q, d_i)$, the relevance score is reestimated as:
\begin{equation*}
    \hat{r}(q, d_i) = \frac{1}{|q|} \sum_{j=1}^{|q|} -\text{log } p(q_j \mid d_i, q_{<j}, I),
\end{equation*}
where $q_j$ denotes the $j$-th token of $q$, $|q|$ denotes the length of $q$, $q_{<j}$ denotes the first $(j-1)$ tokens of $q$, and $I$ represents an instruction.
Note that the language model does not actually perform generation, as we are only estimating the joint probability since the actual query $q$ is given.
Therefore, we can directly employ pre-trained mLMs, without requiring any instruction tuning.
In our framework, we employ the prefix ``\textit{Based on the passage, please write a question in {L}}'' for reranking.

This relevance score can be interpreted as the negative log-likelihood of the mLM generating the query $q$ given the document $d_i$.
Intuitively, the more relevant $d_i$ is to $q$, the more likely the mLM will generate $q$.
Thus, we leverage this property to rerank multilingual passages, even though the mLM is pre-trained without any ranking supervision.
Notably, while this step does not require any paired data, we need a set of multilingual queries, which is comparatively easier to collect than query-document pairs.

\subsection{Knowledge-Distilled Retriever Training}
Previous work has demonstrated that distilling knowledge from a strong reranker can significantly enhance the performance of the retriever~\cite{rosa2022no,li-etal-2022-learning-cross}. In the second stage, we employ the mLM reranker from the first stage as the teacher model to improve the performance of the
e performance of the dense retriever.
We initialize the student model with the multilingual retriever used in the first stage and train it to mimic the outputs of the teacher model by minimizing the Kullback-Leibler (KL) divergence.

Specifically, the relevance of a document $d_i$ to a query $q$ predicted by the student model can be defined as:
\begin{equation*}
    P(d_i \mid q) = \frac{\text{exp} (r(q, d_i))}{\sum_{d_j \in \mathcal{D}_{\mathcal{B}}} \text{exp} (r(q, d_j))},
\end{equation*}
where $\mathcal{D}_{\mathcal{B}}$ denotes the documents in the current batch.
Similarly, the relevance predicted by the teacher model can be defined as:
\begin{equation*}
    \hat{P}(d_i \mid q) = \frac{\text{exp} (\hat{r}(q, d_i) / \tau)}{\sum_{d_j \in \mathcal{D}_{\mathcal{B}}} \text{exp} (\hat{r}(q, d_j) / \tau)},
\end{equation*}
where $\tau$ is the temperature parameter for controlling the sharpness of the distribution.
Finally, the loss function is the KL divergence between two distributions:
\begin{equation*}
    \mathcal{L} = \frac{1}{|\mathcal{B}|} \sum_{q \in \mathcal{B}} \text{KL}(\hat{P}(d \mid q) \parallel P(d \mid q)),
\end{equation*}
where $|\mathcal{B}|$ denotes the size of the batch.
Note that we do not convert rankings into hard labels as done in previous work, where only the top-ranked passage is labeled as positive and the rest are treated as hard negatives. The prior approach disregards the fine-grained scores of the negatively labeled documents, potentially leading to suboptimal knowledge transfer. Instead, we use KL loss to enable the retriever to learn the predicted distribution of the reranker, which we observed improves retrieval performance.

In the retriever training process, in-batch negative examples play a critical role in dense retrieval performance, enabling larger batch sizes while remaining efficient~\cite{karpukhin-etal-2020-dense}. We incorporate this technique in our knowledge distillation process by considering documents from other queries in the same batch as in-batch negatives. The scores of the in-batch negatives are set to a very small number, effectively zeroing their probability after the softmax operation. Specifically, with a batch size of $b$ and $n$ documents per query, each query has $n$ associated reranking scores and $n \times (b-1)$ negative documents.

\subsection{Iterative Training}
To prevent overfitting on the same top-k passages and optimize the retriever's performance, we introduce an iterative training approach.
In each iteration, we use the trained retriever to build an index, retrieve the top-k documents, and perform unsupervised multilingual reranking. We then fine-tune the trained retriever using knowledge-distilled retriever training.
The fine-tuned retriever becomes the retriever for the next iteration. This iterative training allows for refreshing the retrieval index in each iteration, avoiding training solely on the same documents. Notably, in the first iteration where no trained retriever is available, we employ the unsupervised pretrained multilingual retriever, mContriever~\cite{izacard2021contriever}.

\section{Experiments}
Our proposed framework, \textbf{UMR}, can be applied to various multilingual information retrieval tasks, such as \emph{cross-lingual passage retrieval} and \emph{multilingual open-domain question-answering}.
We evaluate our approach on XOR-TYDI QA~\cite{asai-etal-2021-xor}, a popular benchmark for multilingual information retrieval.
We also conduct ablation studies to analyze the impact of different components of our approach.

\subsection{Datasets}
XOR-TYDI QA~\cite{asai-etal-2021-xor} is a multilingual open QA dataset consisting of 7 typologically diverse languages, Arabic, Bengali, Finnish, Japanese, Korean, Russian, and Telugu.
The questions are originally from TYDI QA~\cite{clark-etal-2020-tydi} and posed by native speakers in a naturally information-seeking scenario.
There are two subtasks in XOR-TYDI QA:
\begin{itemize}
    \item \textbf{XOR-Retrieve} requires a system to retrieve English passages given a query in language $L$ other than English.
The evaluation metrics used are R@2kt and R@5kt, which measure the recall by computing the fraction of the questions for which the minimal answer is contained in the top $\{2000, 5000\}$ tokens retrieved.
    \item \textbf{XOR-Full} requires a system to retrieve either English documents or documents in the query language $L$ in order to generate an answer in $L$.
The answers are annotated by 1) extracting spans from Wikipedia in the same language as the question (in-language) or 2) translating English spans extracted from English Wikipedia to the target language (cross-lingual).
The evaluation metrics used are F1, EM, and BLEU.
Note that since \textbf{UMR} is only responsible for retrieving relevant documents, we use the reader model from CORA to generate an answer given the retrieved documents.
For the multilingual passage collection, we directly use the preprocessed passage collection released by CORA~\cite{asai2021one}, which consists of February 2019 Wikipedia dumps of 13 diverse languages from all XOR-TYDI QA languages.
The collection has 44 million passages.

\end{itemize}

\subsection{Baseline Systems}
\begin{itemize}
    \item \textbf{BM25} retrieves passages from the target language only. We use a BM25-based lexical retriever implemented in CORA~\cite{asai2021one}, which uses the implementation from Pyserini~\cite{Lin_etal_SIGIR2021_Pyserini}. The retrieved passages are fed to a multilingual QA model to extract final answers.
    \item \textbf{MT+DPR} first translates the question into English and retrieves English documents with DPR~\cite{karpukhin-etal-2020-dense}, which is a monolingual retriever.
    \item \textbf{mGenQ} generates multilingual questions with \texttt{mT0}\footnote{TyDi QA is part of mT0's training data, which gives this baseline a slight advantage.}, a multilingual instruction-tuned language model. The generated questions are used to train a multilingual retriever. We generate the same amount of questions as the training set of XOR-Retrieve for each language.
    \item \textbf{mDPR}\cite{asai-etal-2021-xor} is a supervised multilingual retriever based on the popular DPR model. It is initialized from mBERT and trained on the training set of XOR-Retrieve and NaturalQuestions~\cite{kwiatkowski-etal-2019-natural}.
    \item \textbf{CORA}~\cite{asai2021one} consists of mDPR and mGEN, which follows the \textit{retrieve-and-generate} recipe. The models are trained on the training set of XOR-Full with iterative data mining.
    \item \textbf{Sentri+mFiD}~\cite{sorokin-etal-2022-ask} is the state-of-the-art system of XOR-Full, which utilizes multilingual translations of the training set and self-training.
\end{itemize}

\begin{table*}[t]
\centering
\resizebox{\linewidth}{!}{
\begin{tabular}{l|ccccccc|c|ccccccc|c}
\toprule
\multirow{2}{*}{\bf Model} & \multicolumn{8}{c}{\bf R@2kt} & \multicolumn{8}{|c}{\bf R@5kt} \\
& Ar & Bn & Fi & Ja & Ko & Ru & Te & Avg & Ar & Bn & Fi & Ja & Ko & Ru & Te & Avg \\
\midrule
\multicolumn{17}{l}{\it Supervised} \\
mDPR & 41.2 & 43.9 & 50.3 & 29.1 & 34.5 & 35.3 & 37.2 & 38.8 & 50.4 & 57.7 & 58.9 & 37.3 & 42.8 & 44.0 & 44.9 & 48.0 \\
MT+DPR & 48.3 & 54.4 & 56.7 & 41.8 & 39.4 & 39.6 & 18.7 & 42.7 & 52.5 & 63.2 & 65.9 & 52.1 & 46.5 & 47.3 & 22.7 & 50.0 \\
\midrule
\multicolumn{17}{l}{\it Unupervised} \\
UMR & 36.7 & 33.6 & 51.6 & 33.2 & 38.3 & 37.2 & 35.8 & 38.1 & 45.0 & 48.8 & 61.9 & 43.4 & 47.3 & 46.9 & 44.4 & 48.2 \\
\bottomrule
\end{tabular}
}
\caption{Performance on XOR-Retrieve test set (\%).}
\label{tab:xor_retrieve}
\end{table*}

\begin{table*}[t]
\centering
\begin{tabular}{l|ccccccc|ccc}
\toprule
\multirow{2}{*}{\bf Model} & \multicolumn{7}{c}{\bf Target Language F1} & \multicolumn{3}{|c}{\bf Macro Average} \\
& Ar & Bn & Fi & Ja & Ko & Ru & Te & F1 & EM & BLEU \\
\midrule
\multicolumn{11}{l}{\it Supervised} \\
MT + DPR & ~~7.6 & ~~5.9 & 16.2 & ~~9.0 & ~~5.3 & ~~5.5 & ~~0.8 & ~~7.2 & ~~3.3 & ~~6.3 \\
CORA & 59.8 & 40.4 & 42.2 & 44.5 & 27.1 & 45.9 & 44.7 & 43.5 & 33.5 & 31.1 \\
Sentri + mFiD & - & - & - & - & - & - & - & 46.2 & 39.0 & 33.7 \\
\midrule
\multicolumn{11}{l}{\it Unsupervised} \\
BM25 & 31.1 & 21.9 & 21.4 & 12.4 & 12.1 & 17.7 & – & – & – & – \\
UMR + CORA Reader & 59.8 & 41.0 & 41.4 & 44.3 & 30.4 & 46.4 & 50.9 & 44.9 & 34.7 & 32.5 \\
\bottomrule
\end{tabular}
\caption{Performance on XOR-Full test set (\%).}
\label{tab:xor_full}
\end{table*}

\subsection{Implementation Details}
For the reranking stage, we retrieve top-100 documents with the trained retriever using a highly-efficient vector search engine, faiss~\cite{douze2024faiss}.
All top-100 documents are reranked by the language modeling-adapted variant of \textbf{mt5-xl}, which has 3 billion parameters~\cite{xue-etal-2021-mt5}.
Note that it is neither fine-tuned on supervised data nor instruction-tuned.

For the knowledge distillation stage, we use \textbf{mContriever} as the initial retriever~\cite{izacard2021contriever}.
In order to reduce memory consumption, we employ the gradient cache technique~\cite{gao-etal-2021-scaling}.
All experiments are conducted on 4xNVIDIA V100 GPUs.
Detailed hyperparameters for training retrievers are shown in Appendix~\ref{sec:parameters}.
We run two iterations of iterative training.

\subsection{Main Results}
\subsubsection{XOR-Retrieve}
The experimental results on the test set of XOR-Retrieve are shown in Table~\ref{tab:xor_retrieve}.
Compared to the supervised baseline mDPR, our proposed \textbf{UMR} achieves comparable or even slightly better performance (48.0\% vs. 48.2\%) despite not using any paired data.
This demonstrates the effectiveness of utilizing mLM for generative pseudo labeling, providing supervision of similar quality compared to human annotation.
The results for each language show that \textbf{UMR} underperforms mDPR significantly in Arabic (Ar) and Bengali (Bn) while achieving comparable or superior performance in other languages.

\subsubsection{XOR-Full}
The experimental results on the test set of XOR-Full are shown in Table~\ref{tab:xor_full}.
Our proposed \textbf{UMR} outperforms a strong supervised baseline CORA and only slightly underperforms the state-of-the-art system Sentri+mFiD.
This result further demonstrates the effectiveness of our proposed method, which requires neither paired data nor query translations.
The performance could be further improved by combining \textbf{UMR} with mFiD, which was shown to be very crucial to the state-of-the-art performance of Sentri~\cite{sorokin-etal-2022-ask}.
Results for each language show that \textbf{UMR} outperforms CORA significantly in Telugu while achieving similar performance in other languages.

\begin{table}[t]
\centering
\begin{tabular}{l|cc}
\toprule
 & \bf R@2kt & \bf R@5kt \\
\midrule
mDPR & 40.50 & 50.20 \\
mGenQ & 29.08 & 38.67 \\
\midrule
mContriever & 25.50	& 35.06 \\
~~~~~+ rerank & 34.24 & 41.88 \\
UMR (iter=1) & 41.23 & 51.50 \\
UMR (iter=2) & 41.68 & 51.94 \\
~~~~~+ rerank & 42.34 & 52.36  \\
\bottomrule
\end{tabular}
\caption{Performance of unsupervised multilingual reranking on XOR-Retrieve dev set (\%). We conduct analyses on the dev set as the test set is not publicly available.}
\label{tab:analysis}
\end{table}

\begin{table}[t]
\centering
\begin{tabular}{l|cc}
\toprule
& \bf R@2kt & \bf R@5kt \\
\midrule
UMR (iter=1) & 41.23 & 51.50 \\
~~~~~- in-batch negative & 39.56 & 49.41 \\
\bottomrule
\end{tabular}
\caption{Performance on XOR-Retrieve dev set with or without using in-batch negatives (\%).}
\label{tab:inbatch}
\end{table}

\begin{table}[t]
\centering
\begin{tabular}{c|cc}
\toprule
Temperature & \bf R@2kt & \bf R@5kt \\
\midrule
1 & 29.58 & 38.82 \\
0.1	& 37.38 & 46.70 \\
0.04 & 37.12 & 46.55 \\
0.02 & 38.43 & 46.45 \\
\bottomrule
\end{tabular}
\caption{Performance on XOR-Retrieve dev set when varying the value of temperature (\%).}
\label{tab:temperature}
\end{table}

\begin{table}[t]
\centering
\begin{tabular}{c|cc}
\toprule
Batch size & \bf R@2kt & \bf R@5kt \\
\midrule
4 & 36.45 & 46.02 \\
8 & 38.94 & 49.38 \\
16 & 40.07 & 50.30 \\
32 & 40.41 & 50.48 \\
\bottomrule
\end{tabular}
\caption{Performance on XOR-Retrieve dev set when varying the value of batch size (\%).}
\label{tab:batch_size}
\end{table}

\begin{table*}[ht]
\centering
\resizebox{\linewidth}{!}{
\begin{tabular}{l|ccc|ccc|ccc}
\toprule
 & \multicolumn{3}{c|}{\bf English Answers Only} & \multicolumn{3}{c|}{\bf Target Language Answers} & \multicolumn{3}{c}{\bf All} \\
 & Top-1 & Top-5 & Top-20 & Top-1 & Top-5 & Top-20 & Top-1 & Top-5 & Top-20 \\
\midrule
\multicolumn{10}{l}{\it Supervised} \\
CORA & 10.8 & 26.9 & 41.8 & 37.0 & 55.0 & 64.9 & 27.1 & 45.7 & 58.1 \\
\midrule
\multicolumn{10}{l}{\it Unsupervised}\\
mContriever & ~~3.2 & ~~7.7 & 13.3 & 18.9 & 40.1 & 56.4 & 14.5 & 31.2 & 45.4 \\
mContriever+rerank & ~~4.4 & ~~9.4 & 15.1 & 29.1 & 50.1 & 61.5 & 20.5 & 37.5 & 49.1 \\
UMR (iter=1) & ~~5.2 & 10.8 & 18.1 & 27.7 & 48.6 & 64.6 & 20.2 & 37.6 & 52.1 \\
UMR (iter=2) & ~~4.7 & 11.4 & 17.9 & 26.2 & 49.2 & 64.6 & 19.1 & 38.5 & 52.1 \\
\bottomrule
\end{tabular}
}
\caption{Retrieval performance on XOR-Full dev set (\%).}
\label{tab:xor_full_retrieve}
\end{table*}

\section{Analysis and Discussion}
In this section, we conduct analytical experiments on the dev set of XOR-Retrieve and XOR-Full since the test sets are not publicly available.

\subsection{Unsupervised Multilingual Reranking}
We conduct an analysis to validate the effectiveness of the unsupervised multilingual reranking stage.
As shown in Table~\ref{tab:analysis}, reranking improves the unsupervised retriever mContriever significantly, improving the result from 25.50 to 34.24 in terms of R@2kt.
This demonstrates that our unsupervised multilingual reranking is effective in reranking the results of the first-stage retriever.
We also observe that the performance of \textbf{UMR} converges after two iterations.
This could be explained by the result of reranking \textbf{UMR} (iter=2), where reranking only achieves a slight improvement.
Given this result, we believe that the performance of \textbf{UMR} is bounded by the reranker.
Future work could explore using more powerful or instruction-tuned LLM and developing superior reranking methods.

\subsection{Question Generation}
Previous work has shown that training a multilingual question generator for generating multilingual questions can improve the performance of multilingual retrieval~\cite{ren-etal-2022-empowering}.
We aim to examine whether this method is feasible in an unsupervised scenario.
We perform multilingual question generation via prompting an instruction-tuned multilingual LLM, \textit{mT0}~\cite{muennighoff2022crosslingual}.
With randomly sampled passages, we generate the same amount of questions as the training set of XOR-Retrieve for each language.
These question-passage pairs are then used to train a multilingual retriever, mGenQ, using the same hyperparameters as mDPR.
The performance of mGenQ is reported in Table~\ref{tab:analysis}.
mGenQ underperforms mDPR and \textbf{UMR} significantly, demonstrating the difficulty of applying question generation to a multilingual scenario where there is no training data.
We manually examine the generated questions and find that roughly half of the questions are either nonsensical or not in the desired language.
Future work could explore effective methods for unsupervised or few-shot multilingual question generation.

\subsection{In-batch Negative}
We conduct an ablation study to validate the effectiveness of the in-batch negative examples.
The results are shown in Table~\ref{tab:inbatch}.
Removing in-batch negatives results in a slight degradation in performance, which is less pronounced compared to supervised dense retrieval methods.
This could be explained by the fact that we include multiple documents per question with fine-grained scores for training, which already includes distinguishing between relevant documents and hard negatives.

\subsection{Effect of Hyperparameters}
Dense retrievers are known to be sensitive to hyperparameters, e.g., batch size.
In this analysis, we examine how different hyperparameters affect the performance of \textbf{UMR}.

\subsubsection{Batch Size}
Training dense retrievers requires a larger batch size.
The results of varying batch sizes are shown in Table~\ref{tab:batch_size}.
When the batch size is under 16, we observe significant degradation in performance.
Hence, in our experiments, we set the batch size to 16.
Note that in our training framework, each question is associated with multiple documents.
Therefore, with a batch size of 16 and 16 documents per question, each question is paired with 256 documents in a batch.

\subsubsection{Temperature}
The results of varying temperature values are shown in Table~\ref{tab:temperature}.
We observe that \textbf{UMR} is highly sensitive to the value of temperature.
When the temperature is set to 1, the performance is degraded significantly from 38.43\% to 29.58\% in terms of R@2kt.
We hypothesize that the range of the negative log-likelihood of the reranker is the root cause of this phenomenon since higher temperature results in a more flat distribution, making it harder for the retriever to learn meaningful knowledge.

\subsection{Retrieval Performance on XOR-Full}
In order to evaluate the multilingual retrieval performance where the language of the relevant documents is not known apriori, we examine the retrieval performance on XOR-Full.
Since there is no official evaluation of the retrieval performance, we take the answers from the dev set, where some of the questions have English answers.
We split the questions into two categories: 1) questions with annotated English answers and 2) questions with only answers in the target language.
We evaluate the retrieval performance by checking whether any of the answers are present in the top-k retrieved documents.
The results are shown in Table~\ref{tab:xor_full_retrieve}.

We observe that despite outperforming CORA in downstream question-answering performance, \textbf{UMR} underperforms CORA significantly in terms of retrieval performance.
This underperformance is especially pronounced in Top-1 recall, which aligns with the observation from ART~\cite{sachan2022questions}.
We hypothesize that while unsupervised reranking via estimating conditional probability can provide good supervision, it cannot distinguish the most relevant documents very well.
We also note that since the reader model takes top-20 passages to generate the answer, Top-20 recall should be a better indicator for the downstream QA performance.
This could explain why \textbf{UMR} achieves better QA performance while performing slightly worse in retrieval performance.
In addition, this evaluation only considers the surface form of the answers, which might fail to capture the difference in surface forms.

\section{Conclusion}
In this paper, we propose \textbf{UMR}, the first unsupervised method for training multilingual dense retrievers without any paired data, which leverages the sequence likelihood estimation capability of pretrained multilingual language models.
The proposed framework consists of two stages with iterative training.
Experimental results on XOR-Retrieve and XOR-Full show that our proposed method performs comparable to or even outperforms strong supervised baselines.
Finally, detailed analyses justify the effectiveness of individual components in our proposed \textbf{UMR}.
We also identify that the performance of \textbf{UMR} might be bounded by the reranking performance of mLM.
Hence, future work could explore better unsupervised reranking methods with large language models.

\section*{Limitations}
While this paper demonstrates the promising performance of our fully unsupervised method for multilingual retrieval, it is important to acknowledge its limitations.

First, our approach assumes that the employed multilingual pre-trained language model already understands the languages present in our evaluated datasets. Consequently, the model's ability to estimate relevance for reranking in the first stage (unsupervised multilingual reranking) relies on this assumption. 
However, for low-resource languages that are not adequately covered by the language model, our proposed approach may struggle to achieve satisfactory performance due to inaccurate estimations.
To address this limitation, we plan to conduct experiments on unseen languages in future work and explore alternative approaches, such as language adaptation techniques, to enhance the generalizability across diverse and even previously unseen languages.

It is crucial to address these limitations to ensure the applicability and effectiveness of our method across a wide range of languages, especially those with limited resources.

\section*{Acknowledgements}
We thank the reviewers for their insightful comments. This work was financially supported by
the National Science and Technology Council (NSTC) in Taiwan, under Grants 111-2222-E-002-
013-MY3, 111-2628-E-002-016, and 112-2223-E002-012-MY5 and Google.

\bibliography{anthology,custom}

\appendix

\section{Hyperparameters}
\label{sec:parameters}
The hyperparameters used for knowledge-distilled retriever training are listed in Table~\ref{tab:parameters}

\begin{table}[h]
\centering
\begin{tabular}{l|r}
\toprule
\multicolumn{2}{l}{\bf hyperparameters} \\
\midrule
max sequence length & 256 \\
batch size & 16 \\
gradient accumulation steps & 1 \\
\# docs per question & 16 \\
train epochs & 10 \\
learning rate & 2e-5 \\
optimizer & AdamW \\
temperature $\tau$ & 0.1 \\
\bottomrule
\end{tabular}
\caption{Hyperparameters used in the knowledge distillation stage.}
\label{tab:parameters}
\end{table}

\end{document}